\documentclass[11pt,a4paper]{article}
\usepackage[usenames,dvipsnames]{xcolor}
\usepackage{ranlp2015}
\usepackage{times}
\usepackage{url}
\usepackage{latexsym} 
\usepackage{graphicx}
\usepackage{multicol}
\usepackage[T1]{fontenc}
\usepackage[utf8]{inputenc}
\usepackage[russian,english]{babel}
\usepackage{pgfplots}
\definecolor{babyblueeyes}{rgb}{0.63, 0.79, 0.95}
\definecolor{babypink}{rgb}{0.96, 0.76, 0.76}

\title{Sentiment Analysis in Twitter for Macedonian}

\author{Dame Jovanoski, Veno Pachovski\\
  University American College Skopje\\
  UACS, Macedonia\\
  {\tt \{jovanoski,pachovski\}@uacs.edu.mk   }  \And
  Preslav Nakov\\
  Qatar Computing Research Institute\\
  HBKU, Qatar\\
  {\tt pnakov@qf.org.qa}}

\date{}

\begin{document}
\maketitle
\begin{abstract}
We present work on sentiment analysis in Twitter for Macedonian. As this is pioneering work for this combination of language and genre, we created suitable resources for training and evaluating a system for sentiment analysis of Macedonian tweets.
In particular, we developed a corpus of tweets annotated with tweet-level sentiment polarity (positive, negative, and neutral), as well as with phrase-level sentiment, which we made freely available for research purposes. We further bootstrapped several large-scale sentiment lexicons for Macedonian, 
motivated by previous work for English. The impact of several different pre-processing steps as well as of various features is shown in experiments that represent the first attempt to build a system for sentiment analysis in Twitter for the morphologically rich Macedonian language. Overall, our experimental results show an F$_1$-score of 92.16, which is very strong and is on par with the best results for English, which were achieved in recent SemEval competitions.

\end{abstract}









\section{Introduction}

The increasing popularity of social media services such as Facebook, Twitter and Google+, and the advance of Web 2.0 have enabled users to share information and, as a result, to have influence on the content distributed via these services. The ease of sharing, e.g., directly from a laptop, a tablet or a smart phone, have contributed to the tremendous growth of the content that users share on a daily basis, to the extent that nowadays social networks have no choice but to filter part of the information stream even when it comes from our closest friends.

Naturally, soon this unprecedented abundance of data has attracted business and research interest from various fields including marketing, political science, and social studies, among many others, which are interested in questions like these: \emph{Do people like the new Apple Watch? What do they hate about iPhone6? Do Americans support ObamaCare?
What do Europeans think of Pope's visit to Palestine? How do we recognize the emergence of health problems such as depression?}

Such questions can be answered by studying the sentiment of the opinions people express in social media.
As a result, the interest for sentiment analysis, especially in social media, has grown, further boosted by the needs of various applications such as mining opinions from product reviews, detecting inappropriate content, and many others.

Below we describe the creation of data and the development of a system for sentiment polarity classification in Twitter for Macedonian: positive, negative, neutral. We are inspired by a similar task at SemEval, which is an ongoing series of evaluations of computational semantic analysis systems, composed by multiple challenges such as text similarity, word sense disambiguation, etc. One of the challenges there was on Sentiment Analysis in Twitter, at SemEval 2013-2015 \cite{Semeval2013,rosenthal-EtAl:2014:SemEval,Rosenthal-EtAl:2015:SemEval,LRE:journal:2015}, where over 40 teams participated three years in a row.\footnote{Other related tasks were the Aspect-Based Sentiment Analysis task \cite{Semeval2014task4,pontiki-EtAl:2015:SemEval}, and the task on Sentiment Analysis of Figurative Language in Twitter \cite{ghosh-EtAl:2015:SemEval}.}
Here we follow a similar setup, focusing on message-level sentiment analysis of tweets, but for Macedonian instead of English.
Moreover, while at SemEval the task organizers used Mechanical Turk to do the annotations, where the control for quality is hard (everybody can pretend to know English), our annotations are done by native speakers of Macedonian.

The remainder of the paper is organized as follows: Section~\ref{sec:related} presents some related work.
Sections~\ref{sec:data} and \ref{sec:lexicons} describe the datasets and the various lexicons we created for Macedonian.
Section~\ref{sec:system} gives detail about our system, including the pre-processing steps and the features used.
Section~\ref{sec:experiments} describes our experiments and discusses the results.
Section~\ref{sec:conclusion} concludes with possible directions for future work.

\section{Related Work}
\label{sec:related}


Research in sentiment analysis started in the early 2000s.
Initially, the problem was regarded as standard document classification into topics, e.g., \newcite{Pang:2002:TUS} experimented with various classifiers such as maximum entropy, Na\"{i}ve Bayes and SVM, using standard features such as unigram/bigrams, word counts/present, word position and part-of-speech tagging. Around the same time, other researchers realized the importance of external sentiment lexicons, e.g., \newcite{turney2002thumbs} proposed an unsupervised approach to learn the sentiment orientation of words/phrases: positive vs. negative. Later work studied the linguistic aspects of expressing opinions, evaluations, and speculations \cite{Wiebe:2004:LSL}, the role of context in determining the sentiment orientation \cite{Wilson05}, of deeper linguistic processing such as negation handling \cite{pang2008opinion}, of finer-grained sentiment distinctions \cite{Pang2005}, 
of positional information \cite{raychev-nakov:2009:RANLP09}, etc. Moreover, it was recognized that in many cases, it is crucial to know not just the polariy of the sentiment, but also the topic towards which this sentiment is expressed \cite{Stoyanov:2008:TIF}.

Early sentiment analysis research focused on customer reviews of movies, and later of hotels, phones, laptops, etc. Later, with the emergence of social media, sentiment analysis in Twitter became a hot research topic.
The earliest Twitter sentiment datasets were both small and proprietary, such as the i-sieve corpus \cite{Kouloumpis11}, or relied on noisy labels obtained from emoticons or hashtags.
This situation changed with the emergence of the SemEval task on Sentiment Analysis in Twitter, which ran in 2013-2015 \cite{Semeval2013,rosenthal-EtAl:2014:SemEval,Rosenthal-EtAl:2015:SemEval}. The task created standard datasets of several thousand tweets annotated for sentiment polarity.
Our work here is inspired by that task.


In our experiments below, we focus on Macedonian, for which we only know two publications on sentiment analysis, none of which is about Twitter.

\newcite{gajduk2014opinion} experimented with 800 posts from the Kajgana forum (260 positive, 260 negative, and 280 objective), using SVM and Naïve Bayes classifiers, and features such as bag of words, rules for negation, and stemming. 

\newcite{uzunova2015sentiment} experimented with 400 movie reviews\footnote{There have been also experiments on movie reviews for the closely related Bulgarian language \cite{sentiBG:RANLP15}, but there the objective was to predict user rating, which was addressed as an ordinal regression problem.} (200 positive, and 200 negative; no objective/neutral), and a Na\"{i}ve Bayes classifier, using a small manually annotated sentiment lexicon of unknown size, and various preprocessing techniques such as negation handling and spelling/character translation.
Unfortunately, the datasets and the generated lexicons used in the above work are not publicly available, and/or are also from a different domain. As we are interested in sentiment analysis of Macedonian tweets, we had to build our own datasets.

In addition to preparing a dataset of annotated tweets, we further focus on creating sentiment polarity lexicons for Macedonian. This is because lexicons are crucial for sentiment analysis. As we mentioned above, since the very beginning, researchers have realized that sentiment analysis was quite different from standard document classification \cite{Sebastiani:2002:MLA:survey}, 
and that 
it crucially needed external knowledge in the form of suitable sentiment polarity lexicons. For further detail, see the surveys by \newcite{pang2008opinion} and \newcite{Liu12}.

Until recently, such sentiment polarity lexicons have been manually crafted, and were of small to moderate size, e.g.,
LIWC \cite{pennebaker01},
General Inquirer \cite{inquirer1966computer},
Bing Liu's lexicon \cite{Hu04},
and MPQA \cite{Wilson05},
all have 2000-8000 words.
Early efforts in building them automatically also yielded lexicons of moderate sizes \cite{esuli:lrec2006,swn}.

However, recent results have shown that automatically extracted large-scale lexicons (e.g., up to a million words and phrases) offer important performance advantages, as confirmed at shared tasks on Sentiment Analysis in Twitter at SemEval 2013-2015 \cite{Semeval2013,rosenthal-EtAl:2014:SemEval,Rosenthal-EtAl:2015:SemEval}.

Similar observations were made in the Aspect-Based Sentiment Analysis task, which ran at SemEval 2014-2015 \cite{Semeval2014task4,pontiki-EtAl:2015:SemEval}. In both tasks, the winning systems benefited from building and using massive sentiment polarity lexicons \cite{MohammadKZ2013,Zhu_SemEval2014}. These large-scale automatic lexicons were typically built using bootstrapping, starting with a small seed of, e.g., 50-60 words \cite{MohammadKZ2013}, and sometimes even using just two emoticons.

\section{Data}
\label{sec:data}

During a period of six months from November 2014 to April 2015, we collected about half a million tweet messages. 
In the process, we had to train and use a high-precision Na\"{i}ve Bayes classifier for detecting the language, because the Twitter API often confused Macedonian tweets with Bulgarian or Russian.
From the resulting set of tweets, we created training and testing datasets, which we manually annotated 
at the tweet level\ (using \emph{positive}, \emph{negative}, and \emph{neutral/objective} as labels\footnote{Following \cite{Semeval2013}, we merged \emph{neutral} and \emph{objective} as they are commonly confused by annotators.}). 

The training dataset was annotated by the first author, who is a native speaker of Macedonian.
In addition to tweet-level sentiment, we also annotated the sentiment-bearing words and phrases inside the \emph{training} tweets, in order to obtain a sentiment lexicon.

The testing dataset was only annotated at the tweet level, and for it there was one additional annotator, again a native speaker of Macedonian.
The value of the Cohen's Kappa statistics \cite{cohen1960} for the inter-annotator agreement between the two annotators was 0.41, which corresponds to \emph{moderate} agreement \cite{Landis1977}; this relatively low agreement shows the difficulty of the task. For the final testing dataset, we discarded all tweets on which the annotators disagreed (a total of 474 tweets).

Table~\ref{tab:data} shows the statistics about the training and the testing datasets. We can see that the data is somewhat balanced between positive and negative tweets, but has a relatively smaller proportion of neutral tweets.\footnote{It was previously reported that most tweets are neutral, but this was for English, and for tweets about selected topics \cite{rosenthal-EtAl:2014:SemEval}. We have no topic restriction; more importantly, there is a severe ongoing political crisis in Macedonia, and thus Macedonian tweets were full of emotions.} 

\begin{table}[tbh]
\begin{center}
\scalebox{0.77}{
\begin{tabular}{l*{5}{c}r}
\bf Dataset & \bf Positive & \bf Neutral & \bf Negative & \bf Total \\ \hline
Train & 2,610 (30\%) & 1,280 (15\%) & 4,693 (55\%) & 8,583 \\
Test & 431 (38\%) & 200 (18\%) & 508 (44\%) & 1,139 \\
\hline
\end{tabular}
}
\end{center}
\caption{\label{tab:data} Statistics about the datasets.}
\end{table}

We faced many problems when processing the tweets.
For example, it was hard to distinguish advertisements vs. news vs. ordinary user messages, which is important for sentiment annotations.
Here is an example tweet by a news agency, which should be annotated as neutral/objective:

\begin{center}
\foreignlanguage{russian}{Лицето АБВ е \textbf{убиецот} и \textbf{виновен} за убиството на БЦД.}
\footnote{Translation: \emph{The person ABC is the killer, and he is responsible for the murder of BCD.}}
\end{center}





The above message has good grammatical structure, but in our datasets there are many messages with missing characters, missing words, misspellings and with poor grammatical structure; this is in part what makes the task difficult. 
Here is a sample message with missing words and misspellings:

\begin{center}
\foreignlanguage{russian}{брао бе, ги утепаа с....!!!}
\footnote{Translation: \emph{That's great, they have smashed them with....!!!}}
\end{center}



Non-standard language is another problem. This includes not only slang and words written in a funny way on purpose, but also many dialectal words from different regions of Macedonia that are not used in Standard Macedonian. For example, in the Eastern part of the Republic of Macedonia, there are words with Bulgarian influence, while in the Western part, there are words influenced by Albanian; and there is Serbian influence in the North.
 

Finally, many problems arise due to our using a small dataset for sentiment analysis. This mainly affects the construction of the sentiment lexicons and the reason for this is the distribution of emoticons, hashtags and sentiment words. In particular, if we want to use hashstags or emoticons as seeds to construct sentiment lexicons, we find that very few tweet messages have emoticons or hashtags. Table~\ref{tab:dist} shows the statistics about the distribution of the emoticons and hashtags in the dataset (half a million tweet messages). That is why, in our experiments below, we do not rely much on hashtags for lexicon construction.

\begin{table}[tbh]
\begin{center}
\scalebox{0.80}{
\begin{tabular}{l*{5}{c}r}
\bf Token type & \bf No. of messages \\
\hline
Without emoticons and hashtags & 473,420 \\
With emoticons & 3,635 \\  
With hashtags & 521 \\
\hline
Total & 477,576 \\
\hline
\end{tabular}
}
\end{center}
\caption{\label{tab:dist} Number of tweets in our datasets that contain emoticons and hashtags.}
\end{table}

\section{Sentiment Lexicons}
\label{sec:lexicons}

Sentiment polarity lexicons are key resources for the task of sentiment analysis, and thus we have put special efforts to generate some for Macedonian using various techniques.\footnote{All lexicons presented here are publicly available at \texttt{https://github.com/badc0re/sent-lex}} Typically, a sentiment lexicon is a set of words annotated with positive and negative sentiment. Sometimes there is also a polarity score of that sentiment, e.g., \emph{spectacular} could have positive strength of 0.91, while for \emph{okay} that might be 0.3.

\subsection{Manually-Annotated Lexicon}

As we mentioned above, in the process of annotation of the training dataset, the annotator also marked the sentiment-bearing words and phrases in each tweet, together with their sentiment polarity in that context: positive or negative. 

The phrases for the lexicon were annotated by two annotators, both native speakers of Macedonian. We calculated the Cohen's Kappa statistics \cite{cohen1960} for the inter-annotator agreement, and obtained the score of 0.63, which corresponds to \emph{substantial} agreement \cite{Landis1977}.

We discarded all words with disagreement, a total of 122, and we collected the remaining words and phrases in a lexicon. The lexicon contained 1,088 words (459 positive and 629 negative).

\subsection{Translated Lexicons}

Another way to obtain a sentiment polarity lexicon is by translating a preexisting one from another language. We translated some English manually-crafted lexicons such as Bing Liu's lexicon (2,006 positive and 4,783 negative), and MPQA (2,718 positive and 4,912 negative),
and an automatically extracted Bulgarian lexicon (5,016 positive and 2,415 negative), extracted from a movie reviews website \cite{sentiBG:RANLP15}.
For the translation of the lexicons we used Google Translate, and we further manually corrected the results, removing bad or missing translations.

\subsection{Automatically-Constructed Lexicons}

Sentiment lexicons can also be constructed automatically by using Pointwise Mutual Information as a way to calculate the semantic orientation of a word \cite{turney2002thumbs} or a phrase in a message (text). In sentiment analysis, using the orientation of a word, the positive and the negative score of a word/phrase can be calculated. The semantic orientation can be calculated as follows:

$SO(w) = PMI(w,pos) - PMI(w,neg)$

\noindent where PMI is the pointwise mutual information, and $pos$ and $neg$ are placeholders standing for any of the seed positive and negative terms.

A positive/negative value for $SO(w)$ indicates positive/negative polarity for $w$, and its magnitude shows the corresponding sentiment strength. In turn, $PMI(w,pos)=\frac{P(w,pos)}{P(w)P(pos)}$, where $P(w,pos)$ is the probability to see $w$ with any of the seed positive words in the same tweet,\footnote{Here we explain the method using number of tweets, as this is how we are using it, but \newcite{turney2002thumbs} actually used page hits in the AltaVista search engine.} $P(w)$ is the probability to see $w$ in any tweet, and $P(pos)$ is the probability to see any of the seed positive words in a tweet;  $PMI(w,neg)$ is defined similarly.

Turney's PMI-based approach further serves as the basis for two popular large-scale automatic lexicons for English sentiment analysis in Twitter, initially developed by NRC for their participation in SemEval-2013 \cite{MohammadKZ2013}.
The \emph{Hashtag Sentiment Lexicon} uses as seeds hashtags containing 32 positive and 36 negative words, e.g.,  \texttt{\#happy} and \texttt{\#sad}; it then uses PMI and extracts 775,000 sentiment words from 135 million tweets.
Similarly, the \emph{Sentiment140} lexicon contains 1.6 million sentiment words and phrases, extracted from the same 135 million tweets, but this time using smileys as seed indicators for positive and negative sentiment, e.g.,  \texttt{:)}, \texttt{:-)} and \texttt{:))} serve as positive seeds, and \texttt{:(} and \texttt{:-(} as negative ones.

In our experiments, we used all words from our manually-crafted Macedonian sentiment polarity lexicon above as seeds, and then we mined additional sentiment-bearing words from a set of half a million Macedonian tweets.
The number of tweets we used was much smaller in scale compared to that used in the \emph{Hashtag Sentiment Lexicon} and in the \emph{Sentiment140} lexicon, since there are much less Macedonian tweets (compared to English).

However, we used a much larger seed; as we will see below, this turns out to be a very good idea. We further tried to construct lexicons using words from the translated lexicons as seeds.


\section{System Overview}
\label{sec:system}

The language of our tweet messages is Macedonian, and thus the text processing is a bit different than for English. As many basic tools that are freely available for English do not exist for Macedonian, we had to implement them in order to improve our model's performance. Our system uses logistic regression for classification, where words are weighted using TF.IDF.

\subsection{Preprocessing}

For pre-processing, we applied various algorithms, which we combined in order to achieve better performance. We used Christopher Potts' tokenizer,\footnote{http://sentiment.christopherpotts.net/tokenizing.html} and we had to be careful since we had to extract not only the words but also other tokens such as hashtags, emoticons, user names, etc. The pre-processing of the tweets goes as follows:

\begin{enumerate}
\item \textbf{URL and username removal}: tokens such as URLs and usernames (i.e., tokens starting with \texttt{@}) were removed.
\item \textbf{Stopword removal}: stopwords were filtered out based on a word list (146 words).
\item \textbf{Repeating characters removal}: consecutive character repetitions in a word were removed; also were removed repetitions of a word in the same token, e.g., `\foreignlanguage{russian}{какоооо}' or `\foreignlanguage{russian}{дадада}' (translated in English as `what' and `yes', respectively).
\item \textbf{Negation handling}: negation was addressed using a predefined list of negation tokens, then the prefix  \texttt{NEG\_CONTEXT\_} was attached to the following tokens until a clause-level punctuation mark, in order to annotate it as appearing in a negated context, as suggested in \cite{Pang:2002:TUS}. A list of 45 negative phrases and words was used to signal negation.
\item \textbf{Non-standard to standard word mapping}: non-standard words (slang) were mapped to an appropriate form, according to a manualy crafted predefined list of mappings.
\item \textbf{PoS tagging}: rule-based, using a dictionary.
\item \textbf{Tagging positive/negative words}: positive and negative words were tagged as \texttt{POS} and \texttt{NEG}, using sentiment lexicons.
\item \textbf{Stemming}: rule-based stemming was performed, which removes/replaces some prefixes/suffixes.
\end{enumerate}


In sum, we started the transformation of an input tweet by converting it to lowercase, followed by removal of URLs and user names. We then normalized some words to Standard Macedonian using a dictionary of 173 known word transformations and we further removed stopwords (a list of 146 words). As part of the transformation, we marked the words in a negated context.

We further created a rule-based stemming algorithm with a list of 65 rules for removing/replacing prefixes and suffixes \cite{porter1980algorithm}. We used two groups of rules: 45 rules for affix removal, and 20 rules for affix replacement. Developing a stemmer for Macedonian was challenging as this is a highly inflective language, rich in both inflectional and derivational forms.
For example, here are some of the forms for the word \foreignlanguage{russian}{навреда} (English noun `\emph{insult, offense}', verb `\emph{offend, insult}'):

\begin{center}
\begin{multicols}{2}
\foreignlanguage{russian}{
\noindentнавредам\\
навредат\\
навредата\\
навредеа\\
навредев\\
навредевме\\
навредевте\\
навредел\\
навредела\\
навределе\\
навредело\\
навреден\\
навредена\\
...
}
\end{multicols}
\end{center}

In total, this word can generate over 90 inflected forms; in some cases, this involves a change in the last letter of the stem.


We further performed PoS (part-of-speech) tagging with our own tool based on averaged perceptron trained on MULTEXT-East resources \cite{Erjavec:2012}.
Here is an annotated tweet:

\begin{center}
\foreignlanguage{russian}{\emph{го/PN даваат/VB Глуп/NN и/CC Поглуп/NN на/CC Телма/NN}}\footnote{The translation for this message is: \emph{Dump and Dumper is on Telma.}}
\end{center}

Here are the POS tags used in the above example: (\emph{i})~NN-noun; (\emph{ii})~AV-adverb; (\emph{iii})~VB-verb; (\emph{iv})~AE-adjective; (\emph{v})~PN-pronoun; (\emph{vi})~PN-pronoun; (\emph{vii})~CN-cardinal number; (\emph{viii})~CC-conjunction.

We also developed a lemmatizer based on \emph{approximate fuzzy string matching}. First, we used the \emph{candidate word} (the one we want to lemmatize) to retrieve word lemmata that are similar to it; we then used \emph{Jaro–Winkler} distance and \emph{Levenshtein} distance to calculate a score that will determine whether the word matches closely enough some of the retrieved words. Such techniques have been used by other authors for \emph{record linkage} \cite{cohen2003comparison}.
Finally, as a last step in the transformation, we weighed the words using TF.IDF.

\subsection{Features}

In order to evaluate the impact of the sentiment lexicon, we defined features that are fully or partially dependent on the lexicons.
When using multiple lexicons at the same time, there are separate instances of these features for each lexicon.
Here are the features we used:
 
(\emph{i})~ Unigrams/bigrams: each one is a feature and its value is its TF.IDF score;
(\emph{ii})~ Number of positive words in the tweet;
(\emph{iii})~ Number of negative words in the tweet;
(\emph{iv})~ Ratio of the number of positive words to the total number of sentiment words in the tweet;
(\emph{v})~ Ratio of negative words to the total number of sentiment words in the tweet;
(\emph{vi})~ Sum of the sentiment scores for all dictionary entries found in the tweet;
(\emph{vii})~ Sum of the positive sentiment scores for all dictionary entries found in the tweet;
(\emph{viii})~ Sum of the negative sentiment scores for all dictionary entries found in the tweet;
(\emph{ix-x})~ Number of positive and negative emoticons in the tweet.


For classification, we used logistic regression. Our basic features were  TF.IDF-weighted unigram and bigrams, and also emoticons.
We further included additional features that focus on the positive and negative terms that occur in the tweet together with their scores in the lexicon. In case of two or more lexicons being used together, we had a copy of each feature for each lexicon.

\section{Experiments}
\label{sec:experiments}

Our evaluation setup follows that of the SemEval 2013-2015 task on Sentiment Analysis in Twitter \cite{Semeval2013,rosenthal-EtAl:2014:SemEval,Rosenthal-EtAl:2015:SemEval}, where the systems were evaluated in terms of an F-score that is the average of the F$_1$-score for the positive, and the F$_1$-score for the negative class. Note that, even though implicit, the neutral class still matters in this score.

\begin{table}[h]
\begin{center}
\scalebox{0.80}{
\begin{tabular}{l*{5}{c}r}
\bf Features & \bf  F-score & \bf Diff. \\
\hline 
All & 92.16 &  \\
\hline  
All - stop words & 86.24 & -5.92 \\
All - negation & 87.51 & -4.65 \\ 
All - norm. words to STD. Macedonian & 90.22 & -1.94 \\
All - repeated characters & 91.10 & -1.06 \\
All - stemming & 93.14 & 0.98 \\
All - PoS & 92.01 & -0.15 \\  
\hline
\end{tabular}
}
\end{center}
\caption{\label{tab:preps} The impact of excluding the preprocessing steps one at a time.}
\end{table}

\begin{table}[h]
\begin{center}
\scalebox{0.80}{
\begin{tabular}{l*{5}{c}r}
\bf Features & \bf  F-score & \bf Diff. \\
\hline 
All & 92.16 &  \\
\hline  
All - automatically-constructed lexicons & 72.77 & -19.39 \\
All - our manually-crafted lexicon & 79.32 & -12.84 \\
All - all translated lexicons & 91.89 & -0.27 \\
\hline
\end{tabular}
}
\end{center}
\caption{\label{tab:lexs} The impact of excluding the features derived from the sentiment polarity lexicons.}
\end{table}

Table~\ref{tab:preps} shows the impact of each pre-processing step. The first row shows the results when using all pre-processing steps and all sentiment lexicons. The following rows show the impact of excluding each of the preprocessing steps, one at a time.
We can see that stopword removal and negation handling are most important: excluding each of them yields a five point absolute from in F-score. Normalization to Standard Macedonian turns out to be very important too as excluding it yields a drop of two points absolute. Handling repeating characters and stemming are also important, each yielding one point drop in F-score. However, the impact of using POS tagging is negligible.

Table~\ref{tab:lexs} shows the impact of excluding some of the lexicons.
We can see that our manually-crafted lexicon is quite helpful, contributing 13 points absolute in the overall F-score.
Yet, the bootstrapped lexicons are even more important as excluding them yields a drop of 19 points absolute.




\section{Conclusion and Future Work}
\label{sec:conclusion}

We have presented work on sentiment analysis in Twitter for Macedonian. As this is pioneering work for this combination of language and genre, we created suitable resources for training and evaluating a system for sentiment analysis of Macedonian tweets.
In particular, we developed a corpus of tweets annotated with tweet-level sentiment polarity (positive, negative, and neutral), as well as with phrase-level sentiment, which we made freely available for research purposes. 

We further bootstrapped several large-scale sentiment lexicons for Macedonian, 
motivated by previous work for English. The impact of several different pre-processing steps as well as of various features is shown in experiments that represent the first attempt to build a system for sentiment analysis in Twitter for the morphologically rich Macedonian language. Overall, our experimental results show an F$_1$-score of 92.16, which is very strong and is on par with the best results for English, which were achieved in recent SemEval competitions.

In future work, we are interested in studying the impact of the raw corpus size, e.g., we could only collect half a million tweets for creating lexicons and analyzing/evaluating the system, while \newcite{kiritchenko2014sentiment} built their lexicon on million tweets and evaluated their system on 135 million English tweets. 
Moreover, we are interested not only in quantity but also in quality, i.e., in studying the quality of the individual words and phrases used as seeds. An interesting work in that direction, even though in a different domain and context, is that of \newcite{kozareva2010not}. 
We are further interested in finding alternative ways for defining the sentiment polarity, including degree of positive or negative sentiment, and in evaluating them by constructing polarity lexicons in new ways \cite{severyn-moschitti:2015:NAACL-HLT}. 

More ambitiously, we would like to extend our system to detecting sentiment over a period of time for the purpose of finding trends towards a topic \cite{Semeval2013,rosenthal-EtAl:2014:SemEval,Rosenthal-EtAl:2015:SemEval}, e.g., predicting whether the sentiment is strongly negative, weakly negative, strongly positive, etc. 
We further plan application to other social media services, with the idea of analyzing the sentiment of an online conversation. We would like to see the impact of earlier messages on the sentiment of newer messages, e.g., as in \cite{vanzo-croce-basili:2014:Coling,barroncedeno-EtAl:2015:ACL-IJCNLP,Joty:EMNLP:2015}.
Finally, we are interested in applying our system to help other tasks, e.g., by using sentiment analysis to finding opinion manipulation trolls in Web forums \cite{mihaylov-georgiev-nakov:2015:CoNLL,trolls:RANLP15}.


\section*{Acknowledgments}
We would like to thank the anonymous reviewers for their constructive comments, which have helped us improve the final version of the paper.

\bibliographystyle{acl}
\bibliography{acl2014}  

\end{document}